\documentclass[letterpaper, 10 pt, conference]{ieeeconf}  

\usepackage{xcolor}
\usepackage{caption}
\usepackage{tabularx}
\usepackage{array}
\usepackage{graphicx}
\usepackage{booktabs}
\usepackage{colortbl}
\usepackage{multirow} 
\usepackage{algorithm}
\usepackage{algorithmic}
\usepackage{amsfonts}
\usepackage{amsmath}

\usepackage[hidelinks]{hyperref}

\newcommand{\projectpageurl}{https://tvcache-vla.github.io/}

\IEEEoverridecommandlockouts              
\title{\LARGE \bf
Text-Vision Synergistic Token Caching: A Training-Free Framework for Efficient Vision-Language-Action Inference
}

\author{
Qianer Li$^{1}$,
Chengjie Zhang$^{1}$,
Jingwen Chen$^{1}$,
Zanjia Tong$^{1}$,
Jiyuan Zhang$^{2}$,
and Hong Zhang$^{1*}$%
\thanks{$^{1}$Qianer Li, Chengjie Zhang, Jingwen Chen, Zanjia Tong, and Hong Zhang are with the College of Engineering, Southern University of Science and Technology (SUSTech), Shenzhen 518055, China.}%
\thanks{$^{2}$Jiyuan Zhang is with the Master of Science in Data Science and Ma- chine Learning , Department of Mathematics, Faculty of Science, National University of Singapore, Singapore.}%
\thanks{Emails: Qianer Li (\texttt{12532498@mail.sustech.edu.cn}); Hong Zhang (\texttt{hzhang@sustech.edu.cn}). $^{*}$Corresponding author.}%
}
\begin{document}

\maketitle
\thispagestyle{empty}
\pagestyle{empty}

\begin{abstract}
Vision-Language-Action (VLA) models enable generalizable robotic control but remain computationally expensive. Token caching provides a training-free, plug-and-play acceleration alternative.
However, existing VLA caching does not fully exploit a key inductive bias of VLA models: \textit{text-vision synergy}, wherein textual semantics guide the precise visual grounding of task-relevant regions. In particular, existing designs insufficiently account for head-wise reliability in attention aggregation and layer-wise stability in cache reuse.
To address this, we propose \textbf{T}ext-\textbf{V}ision Synergistic Token \textbf{Caching} (\textbf{TVCache}), a training-free framework for efficient VLA inference.
TVCache filters attention heads based on text-vision information focus to improve task-relevant and physically consistent visual grounding.
Concurrently, we introduce a reuse-layer selection mechanism guided by text-vision entropy differences to avoid caching unstable representations and improve cache resource allocation.
Extensive experiments across four representative VLA models, two simulation benchmarks, and real-world robotic tasks demonstrate the effectiveness and generality of TVCache. At matched token-retention ratios, TVCache consistently improves task success over existing VLA caching with comparable computational cost. On OpenVLA-OFT, it improves average success by up to 14.5 percentage points over VLA-Cache at 12.5\% retention while reducing FLOPs by $2.45\times$ relative to full-token inference.
\end{abstract}

\section{INTRODUCTION}
\label{introduction}  
Vision-Language-Action (VLA) models have demonstrated strong generalization across diverse robotic manipulation tasks by leveraging large-scale multimodal data and transformer architectures \cite{zitkovich2023rt, kim2024openvla, black2024pi_0}. Recent advances, such as OpenVLA-OFT \cite{kim2025fine}, BitVLA \cite{wang2025bitvla}, VLA-Adapter \cite{wang2025vla}, and $\pi_{0.5}$ \cite{intelligence2025pi_}, further improve adaptation efficiency, computational efficiency, and long-horizon control.

Despite this progress, real-time VLA inference remains costly due to repeated processing of dense visual tokens. Existing acceleration methods, including lightweighting \cite{yang2025efficientvla, ni2025swiftvla, jiang2026fast}, quantization \cite{zheng2026dyq, xu2026qvla}, and early exits \cite{yue2024deer, song2025ceed, hu2025deead}, often require retraining or architectural changes. Token caching \cite{xu2025vla} provides a complementary, training-free alternative by exploiting temporal redundancy across consecutive observations and reusing previously computed key-value (KV) representations. Importantly, caching is largely orthogonal to model-level acceleration and can therefore be combined with lightweighting, quantization, or early-exit techniques for further efficiency gains. 

However, existing VLA caching does not fully exploit a key inductive bias of VLA models: \textit{text-vision synergy}, wherein textual semantics guide the precise visual grounding of task-relevant regions. In particular, existing designs insufficiently account for head-wise reliability during attention aggregation and layer-wise stability during cache reuse \cite{xu2025vla}. We argue that exploiting this synergy is critical not only for identifying \emph{which} visual tokens should be recomputed, but also for determining \emph{where} cached representations should be reused across network depth.

\begin{figure*}[t] 
    \centering  
    \includegraphics[width=0.95\textwidth]{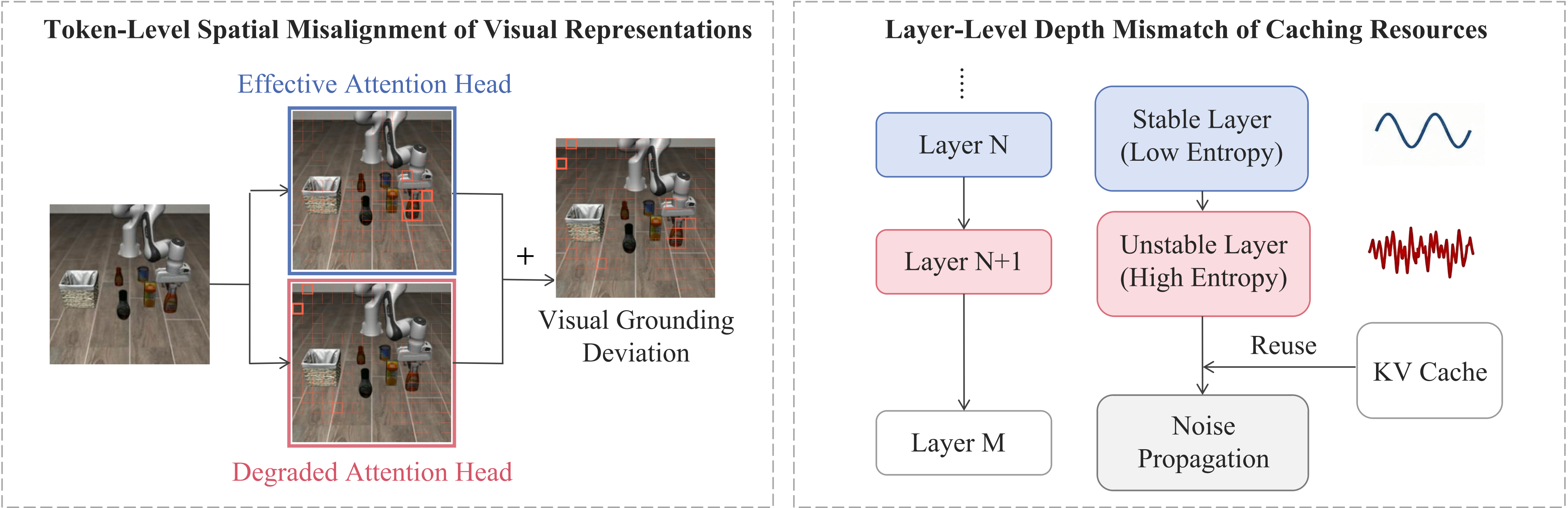}
    \caption{Two bottlenecks of existing VLA token caching. \textbf{(Left)} Token-level spatial misalignment: indiscriminate attention-head aggregation introduces task-irrelevant responses and degrades visual grounding. \textbf{(Right)} Layer-level depth mismatch: reusing representations at unstable layers can propagate noisy cross-modal features, whereas delaying reuse after representations stabilize incurs redundant computation.}
    \label{fig:introduction}
    \vspace{-0.3cm}
\end{figure*}

As illustrated in Fig.~\ref{fig:introduction}, insufficient exploitation of text-vision synergy manifests as two bottlenecks.
First, at the token level, indiscriminately aggregating all attention heads mixes informative task-conditioned responses with diffuse or task-irrelevant attention. The resulting relevance scores can therefore be biased toward background regions, causing token-level spatial misalignment and reducing the reliability of task-relevant token selection.
Second, at the layer level, reuse policies without explicit stable-layer selection overlook the fact that cross-modal representations exhibit different degrees of stability across network depth.
Reusing representations at unstable, high-entropy layers can propagate unreliable features, whereas postponing reuse after representations have already stabilized introduces unnecessary computation. We refer to this inefficiency as layer-level depth mismatch.

To address these limitations, we propose \textbf{T}ext-\textbf{V}ision Synergistic Token \textbf{Caching} (\textbf{TVCache}), a training-free and plug-and-play framework for efficient VLA inference. TVCache exploits text-vision synergy along two complementary dimensions.
\textbf{Token-Level}, TVCache resolves spatial misalignment via an \textbf{attention-head filtering mechanism driven by text-vision information focus}, which we quantify by jointly evaluating semantic response intensity and attention entropy. By dynamically selecting high-quality attention heads based on this evaluation, the mechanism effectively suppresses interference from degraded heads. Consequently, it guides token selection toward task-relevant regions, achieving more precise and physically consistent visual grounding.
\textbf{Layer-Level}, it mitigates depth mismatch using a \textbf{reuse-layer selection mechanism guided by text-vision entropy differences}. By quantifying the stability of layer-wise cross-modal features, it selects low-entropy caching locations while avoiding unstable layers. This mechanism limits the propagation of unreliable features and reduces redundant computation, thereby enabling more efficient cache resource allocation.

Extensive experiments across four representative VLA models \cite{kim2025fine, wang2025bitvla, wang2025vla, intelligence2025pi_}, two simulation benchmarks \cite{liu2023libero, mees2022calvin}, and real-world robotic tasks demonstrate the effectiveness and generality of TVCache. At matched token-retention ratios, TVCache consistently improves task success over existing VLA caching while maintaining comparable computational cost. On OpenVLA-OFT, TVCache improves average success by up to 14.5 percentage points over VLA-Cache, from 69.5\% to 84.0\% at 12.5\% token retention, while reducing FLOPs by $2.45\times$ relative to full-token inference. Our main contributions are:
1) We identify insufficient exploitation of text-vision synergy as a key limitation of existing VLA caching, and propose TVCache to address the resulting token-level and layer-level inefficiencies in a training-free manner.
2) We introduce an attention-head filtering mechanism driven by text-vision information focus that mitigates token-level spatial misalignment and improves task-relevant visual grounding.
3) We develop a text-vision entropy-guided reuse-layer selection mechanism that mitigates layer-level depth mismatch by avoiding unstable reuse locations and reducing redundant computation.

\section{Related Work}
\label{related_work}
\textbf{Vision-Language-Action Models.}
Built upon powerful Large-scale Vision-Language Models (VLMs) \cite{touvron2023llama,  ahmed2025qwen}, Vision-Language-Action (VLA) models achieve remarkable control performance via discretized tokens \cite{kim2024openvla} or policy heads \cite{black2024pi_0}. However, processing dense, high-dimensional visual tokens incurs prohibitive computational overhead, severely bottlenecking their real-time deployment in dynamic physical environments \cite{li2024evaluating}. 

\textbf{Visual Token Pruning for VLMs.}
Visual token pruning reduces inference cost by removing redundant visual tokens \cite{yao2026towards}. Representative methods, such as FastV \cite{chen2024image}, SparseVLM \cite{zhang2024sparsevlm}, and DivPrune \cite{alvar2025divprune}, have demonstrated effective token reduction for general-purpose multimodal models. However, robotic manipulation places stronger demands on preserving fine-grained, task-conditioned visual information, since small interaction regions may directly affect downstream actions. This motivates VLA-specific acceleration mechanisms that explicitly account for language-guided task relevance during token selection.

\textbf{Acceleration Techniques for VLA Models.} 
Existing VLA acceleration methods include lightweighting \cite{yang2025efficientvla, jiang2026fast}, quantization \cite{zheng2026dyq, xu2026qvla}, and early exits \cite{yue2024deer, hu2025deead}. While effective, these approaches typically require retraining, calibration, or architectural modifications. Token caching provides a complementary, training-free alternative by exploiting temporal redundancy and reusing previously computed KV representations \cite{xu2025vla}.
VLA-Cache \cite{xu2025vla} further incorporates text-to-vision attention to protect task-relevant tokens and adapts token reuse across network depth. Nevertheless, its attention aggregation does not explicitly model head-wise reliability, and its reuse policy does not explicitly select caching locations according to layer-wise representation stability. TVCache builds upon this caching paradigm by more fully exploiting \textit{text-vision synergy}: it filters attention heads according to text-vision information focus for task-relevant token selection and selects reuse layers according to text-vision entropy differences. These two mechanisms respectively target token-level spatial misalignment and layer-level depth mismatch while preserving the training-free nature of token caching.

\begin{figure*}[t] 
    \centering  
    \includegraphics[width=0.87\textwidth]{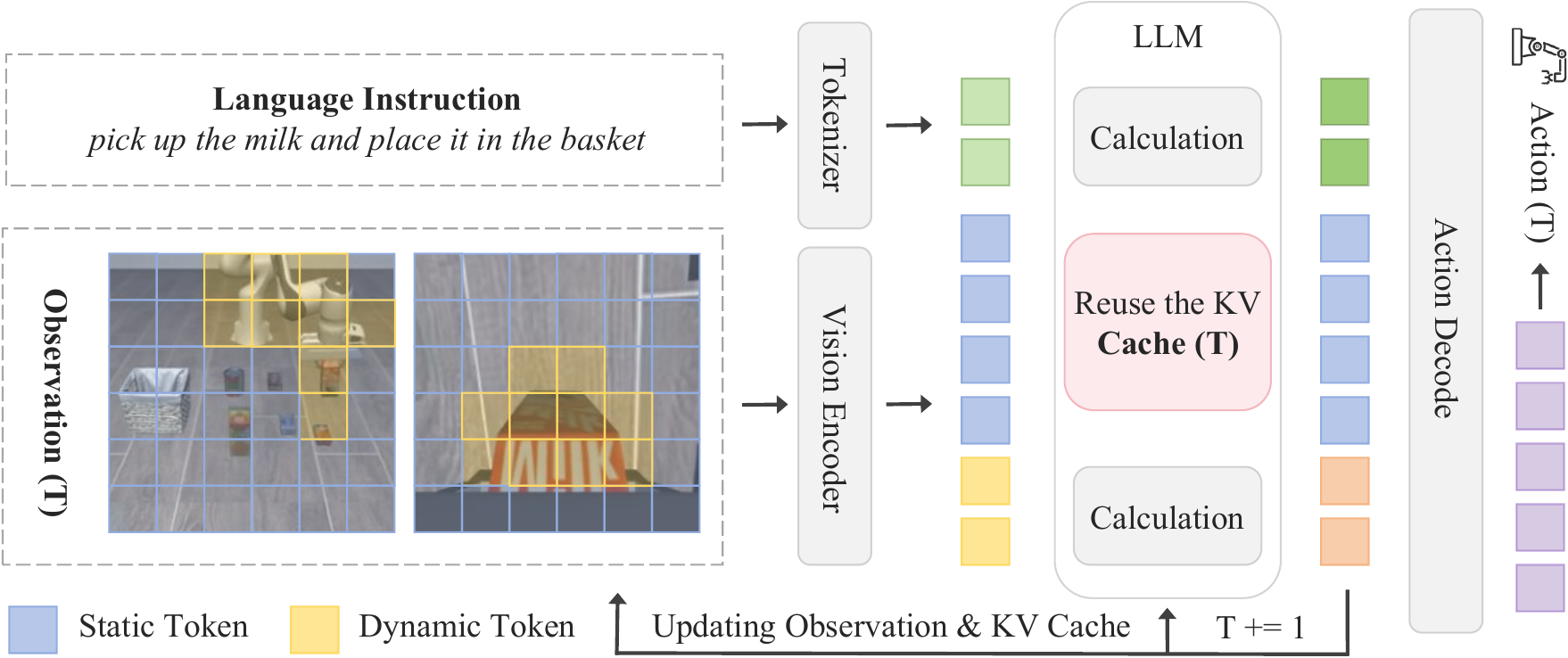}
    \caption{Principle of KV cache reuse in VLA inference. At timestep $T$, visual inputs are decoupled into static (blue) and dynamic (yellow) tokens. By reusing historical caches for invariant instructions and static tokens, the LLM allocates computation exclusively to new dynamic tokens, significantly accelerating autoregressive decoding.}  
    \label{fig:kv_cache_principle}
    \vspace{-0.3cm}
\end{figure*}

\section{Methodology}
\label{sec:methodology}

\subsection{Overview of the TVCache Framework}

\textbf{KV Caching in VLA Inference.}
Continuous robotic manipulation exhibits strong temporal coherence across observations. As shown in Fig.~\ref{fig:kv_cache_principle}, token caching reuses historical KV representations for stable visual regions while recomputing dynamic ones, reducing redundant Transformer computation \cite{xu2025vla}.

\begin{figure*}[t] 
    \centering  
    \includegraphics[width=1.0\textwidth]{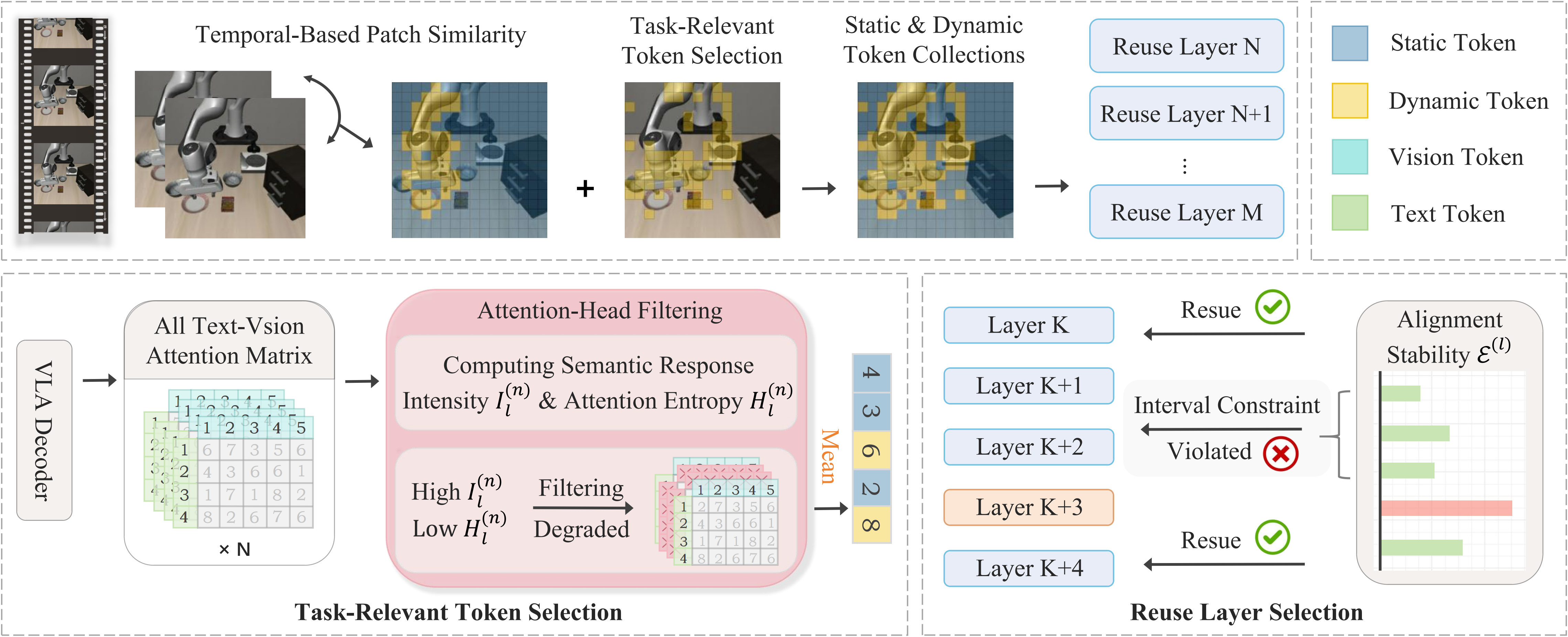}
    \caption{Overview of the proposed framework.}  
    \label{fig:tvkv_cache_principle}
    \vspace{-0.3cm}
\end{figure*}

\textbf{The TVCache Framework.}
Existing VLA caching already exploits temporal similarity and text-conditioned attention for token reuse \cite{xu2025vla}, yet it does not fully account for two reliability factors identified in Sec.~\ref{introduction}: head-wise reliability during task-relevant token estimation and layer-wise stability during cache reuse. TVCache builds upon this caching paradigm and exploits text-vision synergy along two complementary dimensions:
(1) Token-Level, an attention-head filtering mechanism driven by text-vision information focus evaluates individual attention heads before estimating token relevance, reducing interference from diffuse or task-irrelevant responses and mitigating token-level spatial misalignment.
(2) Layer-Level, a reuse-layer selection mechanism guided by text-vision entropy differences identifies stable reuse locations across network depth, mitigating layer-level depth mismatch while reducing redundant computation. The overall framework is illustrated in Fig.~\ref{fig:tvkv_cache_principle}.

\subsection{Temporal Patch Similarity}
Let $p_j^t$ denote the $j$-th visual patch at timestep $t$. We select the top-$k$ patches with highest temporal similarity above threshold $\tau$:
\begin{equation}
P_{\mathrm{static}}
=
\mathrm{Top}\text{-}k
\left(
\left\{
p_j^t
\mid
\mathrm{Sim}(p_j^t,p_j^{t-1}) \ge \tau
\right\}
\right),
\end{equation}
where $\mathrm{Sim}$ denotes cosine similarity, and $k$ is determined by the token-retention budget. The resulting $P_{\mathrm{static}}$ forms the initial cache-reuse candidates \cite{choi2024vid, xu2025vla}. TVCache then further filters them by task relevance.


\subsection{Task-Relevant Token Selection via Attention-Head Filtering}
We next determine which temporally stable tokens remain important to the current instruction. Text-to-vision attention provides a natural signal for this purpose because it reflects how textual semantics are grounded onto visual regions. However, different attention heads exhibit heterogeneous grounding behavior: some focus sharply on interaction-relevant regions, whereas others distribute attention broadly over background or task-irrelevant content. Directly averaging all heads can therefore dilute informative responses and distort token relevance estimation.

To capture text-vision semantic dependencies, we extract the cross-modal attention sub-tensor
$\mathbf{C}_l \in \mathbb{R}^{N_H \times L_t \times L_v}$
from the full attention matrix $\mathbf{A}_l$, where $N_H$, $L_t$, and $L_v$ denote the numbers of attention heads, text tokens, and visual tokens, respectively. Let $c_{i,j}^{(n)}$ denote the attention weight from text token $i$ to visual token $j$ at head $n$.

Rather than treating all heads equally, TVCache explicitly evaluates their text-vision information focus. We characterize this property from two complementary perspectives. First, semantic response intensity measures how strongly a head responds from textual tokens to visual tokens, reflecting its participation in cross-modal interaction. Second, attention entropy measures the spatial concentration of this response: lower entropy indicates that the head focuses its attention on a smaller set of visual regions rather than dispersing it broadly. These two quantities are respectively defined as semantic response intensity $I_l^{(n)}$ and attention entropy $H_l^{(n)}$:
\begin{equation}
I_l^{(n)} = \frac{1}{L_t} \sum_{i=1}^{L_t} \sum_{j=1}^{L_v} c_{i,j}^{(n)}, \quad
H_l^{(n)} = - \frac{1}{L_t} \sum_{i=1}^{L_t} \sum_{j=1}^{L_v} \tilde{c}_{i,j}^{(n)} \log \tilde{c}_{i,j}^{(n)},
\end{equation}
where $\tilde{c}_{i,j}^{(n)}$ represents the attention weight normalized over the visual dimension to form a valid probability distribution. We combine the two signals through the scoring function $\mathcal{S}_{\text{head}}(n) = \alpha I_l^{(n)} - \beta H_l^{(n)}$. Heads with higher scores exhibit stronger text-conditioned responses together with more concentrated spatial attention and are selected to form the head set $\mathcal{M}_l$. 
We set $\alpha=\beta=0.5$ for all experiments without model- or benchmark-specific tuning.


The selected heads then provide a more reliable source for estimating visual-token relevance. Specifically, the task-relevant score $V_j$ is computed as
$V_j = \frac{1}{|\mathcal{M}_l|} \sum_{m \in \mathcal{M}_l} \sum_{i=1}^{L_t} c_{i,j}^{(m)}$.
Tokens exceeding the threshold $\tau_\text{task}$ form the interaction-critical set $P_{\text{task}}$. These tokens are excluded from the temporally stable candidates, yielding the final reusable set $P_{\text{reuse}} = P_{\text{static}} \setminus P_{\text{task}}$.


\subsection{Cache Resource Allocation via Reuse-Layer Selection}
Identifying $P_{\text{reuse}}$ answers \emph{what} visual information can be reused, but reliable caching also requires deciding \emph{where} these representations should be reused across network depth. Cross-modal representations evolve progressively through the Transformer and exhibit different degrees of concentration across layers. Consequently, applying reuse without explicitly considering layer-wise stability may either reuse representations before they become sufficiently reliable or delay reuse after they have already stabilized, resulting in unnecessary computation.

TVCache therefore uses text-vision entropy differences to characterize layer-wise cross-modal stability. Specifically, we compute the Shannon entropy of normalized text-to-vision attention at each layer:
\begin{equation}
\mathcal{E}^{(l)} = \mathcal{H}(\tilde{\mathbf{C}}_l),
\end{equation}
where $\tilde{\mathbf{C}}_l$ is the text-to-vision attention normalized spatially, and $\mathcal{H}(\cdot)$ denotes Shannon entropy. A lower entropy corresponds to a more concentrated text-vision attention distribution and is therefore used as an indicator of a more suitable reuse location. We define the corresponding layer stability score as $R^{(l)} = 1 - \bar{\mathcal{E}}^{(l)}$, where $\bar{\mathcal{E}}^{(l)}$ is the min-max normalized entropy across all layers.
 
Given a cache budget of $K$ layers out of $N_L$ Transformer layers, TVCache selects reuse locations that jointly favor high layer stability while avoiding excessive concentration within a narrow depth range. We therefore formulate reuse-layer selection as the following constrained objective:
\begin{equation}
\begin{aligned}
\mathcal{K}_{\mathrm{reuse}}
=
\underset{
\mathcal{K} \subset \{1,\dots,N_L\}
}{
\arg\max
}
&\quad
\sum_{l \in \mathcal{K}} R^{(l)}
\\
\text{s.t.}
&\quad
|\mathcal{K}| = K,
\\
&\quad
|l_i-l_j| \ge d_{\min},
\quad
\forall\, l_i \neq l_j \in \mathcal{K}.
\end{aligned}
\end{equation}

For inference efficiency, we solve this via an interval-constrained greedy algorithm, sequentially appending layers with the highest $R^{(l)}$ to $\mathcal{K}_{\mathrm{reuse}}$ while strictly satisfying $d_{\min}$.

\textbf{Depth-Aware Caching Proportions.} 
After determining the reuse locations, TVCache further assigns a depth-aware retention proportion $\gamma^{(l)}$ to each selected layer $l \in \mathcal{K}_{\mathrm{reuse}}$. 
Recognizing that shallow layers process redundant local patterns while deeper layers encode dense semantics vulnerable to representation collapse \cite{jeddi2025similarity}, we enforce a monotonically increasing retention schedule. Shallow layers aggressively decimate tokens for efficiency, whereas deeper layers progressively restore them, culminating in strict full retention ($\gamma=1.0$) at the terminal reuse layer.

\textbf{Synergistic Inference Execution.} 
During the forward pass, for each layer $l \in \mathcal{K}_{\mathrm{reuse}}$, the model directly loads the KV states of the token subset $P_{\text{reuse}}$ scaled by $\gamma^{(l)}$, bypassing their self-attention computation. This mechanism optimizes cache resource allocation by eliminating redundant computations in stable layers while preventing noise accumulation across unstable ones.


\begin{table*}[t]
  \caption{Comparison of latency and performance between OpenVLA-OFT and BitVLA evaluated on the LIBERO Benchmark. The task success rate retention ratios (\%) and speedup multipliers ($\times$) are calculated relative to the Vanilla baseline of each respective model at 100\% tokens.}
  \label{tab:performance_metrics_1}
  \centering
  \resizebox{\textwidth}{!}{
  \begin{tabular}{lcccccccc}
  \toprule
  Method & Spatial (\%) & Object (\%) & Goal (\%) & Long (\%) & Avg Success Rate (\%) $\uparrow$ & FLOPs (T) $\downarrow$ & Control Freq. (Hz) $\uparrow$ & Latency (ms) $\downarrow$ \\ 
  \midrule
  
  \rowcolor[HTML]{DDEBF7} 
  \multicolumn{9}{c}{\textbf{OpenVLA-OFT}} \\ \midrule
  
  \rowcolor[HTML]{F2F2F2} \multicolumn{9}{c}{100\% Tokens} \\
  Vanilla   & 99.00 & 98.00 & 97.20 & 94.40 & 97.15 \,{\scriptsize\textcolor{gray}{(100.0\%)}} & 4.0139 \,{\scriptsize\textcolor{gray}{(1.00$\times$)}} & 10.43 \,{\scriptsize\textcolor{gray}{(1.00$\times$)}} & 95.87 \,{\scriptsize\textcolor{gray}{(1.00$\times$)}} \\ \midrule
  
  \rowcolor[HTML]{F2F2F2} \multicolumn{9}{c}{Retain 50\% Tokens} \\
  FastV     & 95.40 & 98.60 & 96.80 & 93.40 & 96.05 \,{\scriptsize\textcolor{gray}{(98.9\%)}} & 2.4134 \,{\scriptsize\textcolor[HTML]{228B22}{(\textbf{1.66$\times$})}} & 11.84 \,{\scriptsize\textcolor[HTML]{228B22}{(\textbf{1.13$\times$})}} & 84.48 \,{\scriptsize\textcolor[HTML]{228B22}{(\textbf{1.13$\times$})}} \\
  SparseVLM & 96.20 & 88.40 & 96.20 & 75.80 & 89.15 \,{\scriptsize\textcolor{gray}{(91.8\%)}} & 2.6809 \,{\scriptsize\textcolor[HTML]{228B22}{(\textbf{1.50$\times$})}} & 11.71 \,{\scriptsize\textcolor[HTML]{228B22}{(\textbf{1.12$\times$})}} & 85.41 \,{\scriptsize\textcolor[HTML]{228B22}{(\textbf{1.12$\times$})}} \\
  DivPrune  & 94.40 & 97.00 & 95.80 & 92.40 & 94.90 \,{\scriptsize\textcolor{gray}{(97.7\%)}} & 2.2921 \,{\scriptsize\textcolor[HTML]{228B22}{(\textbf{1.75$\times$})}} & 11.86 \,{\scriptsize\textcolor[HTML]{228B22}{(\textbf{1.14$\times$})}} & 84.32 \,{\scriptsize\textcolor[HTML]{228B22}{(\textbf{1.14$\times$})}} \\
  VLA-Cache & 98.20 & 98.20 & 97.40 & 93.40 & 96.80 \,{\scriptsize\textcolor{gray}{(99.6\%)}} & 2.6927 \,{\scriptsize\textcolor[HTML]{228B22}{(\textbf{1.49$\times$})}} & 11.72 \,{\scriptsize\textcolor[HTML]{228B22}{(\textbf{1.12$\times$})}} & 85.29 \,{\scriptsize\textcolor[HTML]{228B22}{(\textbf{1.12$\times$})}} \\
  TVCache (Ours)      & \textbf{99.20} & \textbf{98.80} & \textbf{97.80} & \textbf{94.40} & \textbf{97.55} \,{\scriptsize\textcolor{gray}{(100.4\%)}} & 2.6913 \,{\scriptsize\textcolor[HTML]{228B22}{(\textbf{1.49$\times$})}} & 11.76 \,{\scriptsize\textcolor[HTML]{228B22}{(\textbf{1.13$\times$})}} & 85.06 \,{\scriptsize\textcolor[HTML]{228B22}{(\textbf{1.13$\times$})}} \\ \midrule
  
  \rowcolor[HTML]{F2F2F2} \multicolumn{9}{c}{Retain 25\% Tokens} \\
  FastV     & 93.20 & 88.40 & 97.00 & 90.60 & 92.30 \,{\scriptsize\textcolor{gray}{(95.0\%)}} & 1.6453 \,{\scriptsize\textcolor[HTML]{228B22}{(\textbf{2.44$\times$})}} & 14.74 \,{\scriptsize\textcolor[HTML]{228B22}{(\textbf{1.41$\times$})}} & 67.83 \,{\scriptsize\textcolor[HTML]{228B22}{(\textbf{1.41$\times$})}} \\
  SparseVLM & 97.40 & 88.40 & 92.20 & 73.40 & 87.85 \,{\scriptsize\textcolor{gray}{(90.4\%)}} & 1.9351 \,{\scriptsize\textcolor[HTML]{228B22}{(\textbf{2.07$\times$})}} & 14.45 \,{\scriptsize\textcolor[HTML]{228B22}{(\textbf{1.39$\times$})}} & 69.22 \,{\scriptsize\textcolor[HTML]{228B22}{(\textbf{1.39$\times$})}} \\
  DivPrune  & 92.20 & 89.60 & 94.20 & \textbf{91.60} & 91.90 \,{\scriptsize\textcolor{gray}{(94.6\%)}} & 1.4442 \,{\scriptsize\textcolor[HTML]{228B22}{(\textbf{2.78$\times$})}} & 15.08 \,{\scriptsize\textcolor[HTML]{228B22}{(\textbf{1.45$\times$})}} & 66.31 \,{\scriptsize\textcolor[HTML]{228B22}{(\textbf{1.45$\times$})}} \\
  VLA-Cache & 96.40 & 84.40 & 94.00 & 76.60 & 87.85 \,{\scriptsize\textcolor{gray}{(90.4\%)}} & 2.0575 \,{\scriptsize\textcolor[HTML]{228B22}{(\textbf{1.95$\times$})}} & 14.60 \,{\scriptsize\textcolor[HTML]{228B22}{(\textbf{1.40$\times$})}} & 68.51 \,{\scriptsize\textcolor[HTML]{228B22}{(\textbf{1.40$\times$})}} \\
  TVCache (Ours)      & \textbf{98.20} & \textbf{96.00} & \textbf{97.80} & 90.20 & \textbf{95.55} \,{\scriptsize\textcolor{gray}{(98.4\%)}} & 2.0569 \,{\scriptsize\textcolor[HTML]{228B22}{(\textbf{1.95$\times$})}} & 14.64 \,{\scriptsize\textcolor[HTML]{228B22}{(\textbf{1.40$\times$})}} & 68.29 \,{\scriptsize\textcolor[HTML]{228B22}{(\textbf{1.40$\times$})}} \\ \midrule
  
  \rowcolor[HTML]{F2F2F2} \multicolumn{9}{c}{Retain 12.5\% Tokens} \\
  FastV     & 91.40 & 48.80 & \textbf{94.80} & 44.00 & 69.75 \,{\scriptsize\textcolor{gray}{(71.8\%)}} & 1.2641 \,{\scriptsize\textcolor[HTML]{228B22}{(\textbf{3.18$\times$})}} & 15.33 \,{\scriptsize\textcolor[HTML]{228B22}{(\textbf{1.47$\times$})}} & 65.22 \,{\scriptsize\textcolor[HTML]{228B22}{(\textbf{1.47$\times$})}} \\
  SparseVLM & 88.40 & 56.80 & 83.20 & 44.60 & 68.25 \,{\scriptsize\textcolor{gray}{(70.3\%)}} & 1.4024 \,{\scriptsize\textcolor[HTML]{228B22}{(\textbf{2.86$\times$})}} & 14.85 \,{\scriptsize\textcolor[HTML]{228B22}{(\textbf{1.42$\times$})}} & 67.34 \,{\scriptsize\textcolor[HTML]{228B22}{(\textbf{1.42$\times$})}} \\
  DivPrune  & 87.60 & 49.20 & 86.40 & 57.00 & 70.05 \,{\scriptsize\textcolor{gray}{(72.1\%)}} & 1.1232 \,{\scriptsize\textcolor[HTML]{228B22}{(\textbf{3.57$\times$})}} & 15.46 \,{\scriptsize\textcolor[HTML]{228B22}{(\textbf{1.48$\times$})}} & 64.67 \,{\scriptsize\textcolor[HTML]{228B22}{(\textbf{1.48$\times$})}} \\
  VLA-Cache & 83.60 & 62.00 & 82.80 & 49.60 & 69.50 \,{\scriptsize\textcolor{gray}{(71.5\%)}} & 1.6463 \,{\scriptsize\textcolor[HTML]{228B22}{(\textbf{2.44$\times$})}} & 15.03 \,{\scriptsize\textcolor[HTML]{228B22}{(\textbf{1.44$\times$})}} & 66.54 \,{\scriptsize\textcolor[HTML]{228B22}{(\textbf{1.44$\times$})}} \\
  TVCache (Ours)      & \textbf{96.80} & \textbf{84.80} & 94.40 & \textbf{60.00} & \textbf{84.00} \,{\scriptsize\textcolor{gray}{(86.5\%)}} & 1.6414 \,{\scriptsize\textcolor[HTML]{228B22}{(\textbf{2.45$\times$})}} & 15.04 \,{\scriptsize\textcolor[HTML]{228B22}{(\textbf{1.44$\times$})}} & 66.51 \,{\scriptsize\textcolor[HTML]{228B22}{(\textbf{1.44$\times$})}} \\ 
  \midrule\midrule
  
  \rowcolor[HTML]{DDEBF7} 
  \multicolumn{9}{c}{\textbf{BitVLA}} \\ \midrule
  
  \rowcolor[HTML]{F2F2F2} \multicolumn{9}{c}{100\% Tokens} \\
  Vanilla   & 97.60 & 99.40 & 91.40 & 87.20 & 93.90 \,{\scriptsize\textcolor{gray}{(100.0\%)}} & 1.5571 \,{\scriptsize\textcolor{gray}{(1.00$\times$)}} & 5.53 \,{\scriptsize\textcolor{gray}{(1.00$\times$)}} & 180.89 \,{\scriptsize\textcolor{gray}{(1.00$\times$)}} \\ \midrule
  
  \rowcolor[HTML]{F2F2F2} \multicolumn{9}{c}{Retain 50\% Tokens} \\
  FastV     & \textbf{98.20} & \textbf{99.40} & 93.40 & 83.80 & 93.70 \,{\scriptsize\textcolor{gray}{(99.8\%)}} & 0.9952 \,{\scriptsize\textcolor[HTML]{228B22}{(\textbf{1.56$\times$})}} & 5.86 \,{\scriptsize\textcolor[HTML]{228B22}{(\textbf{1.06$\times$})}} & 170.76 \,{\scriptsize\textcolor[HTML]{228B22}{(\textbf{1.06$\times$})}} \\
  SparseVLM & 95.40 & 97.40 & 87.00 & 83.60 & 90.85 \,{\scriptsize\textcolor{gray}{(96.8\%)}} & 1.0807 \,{\scriptsize\textcolor[HTML]{228B22}{(\textbf{1.44$\times$})}} & 5.76 \,{\scriptsize\textcolor[HTML]{228B22}{(\textbf{1.04$\times$})}} & 173.65 \,{\scriptsize\textcolor[HTML]{228B22}{(\textbf{1.04$\times$})}} \\
  DivPrune  & 94.20 & 97.00 & 82.60 & 74.20 & 87.00 \,{\scriptsize\textcolor{gray}{(92.7\%)}} & 0.9087 \,{\scriptsize\textcolor[HTML]{228B22}{(\textbf{1.71$\times$})}} & 5.87 \,{\scriptsize\textcolor[HTML]{228B22}{(\textbf{1.06$\times$})}} & 170.22 \,{\scriptsize\textcolor[HTML]{228B22}{(\textbf{1.06$\times$})}} \\
  VLA-Cache & 97.20 & 98.80 & 93.60 & 86.00 & 93.90 \,{\scriptsize\textcolor{gray}{(100.0\%)}} & 1.0941 \,{\scriptsize\textcolor[HTML]{228B22}{(\textbf{1.42$\times$})}} & 5.76 \,{\scriptsize\textcolor[HTML]{228B22}{(\textbf{1.04$\times$})}} & 173.58 \,{\scriptsize\textcolor[HTML]{228B22}{(\textbf{1.04$\times$})}} \\
  TVCache (Ours)      & 98.00 & \textbf{99.40} & \textbf{95.00} & \textbf{87.20} & \textbf{94.90} \,{\scriptsize\textcolor{gray}{(101.1\%)}} & 1.0886 \,{\scriptsize\textcolor[HTML]{228B22}{(\textbf{1.43$\times$})}} & 5.80 \,{\scriptsize\textcolor[HTML]{228B22}{(\textbf{1.05$\times$})}} & 172.34 \,{\scriptsize\textcolor[HTML]{228B22}{(\textbf{1.05$\times$})}} \\ \midrule
  
  \rowcolor[HTML]{F2F2F2} \multicolumn{9}{c}{Retain 25\% Tokens} \\
  FastV     & 97.40 & \textbf{99.20} & 86.60 & 78.40 & 90.40 \,{\scriptsize\textcolor{gray}{(96.3\%)}} & 0.7207 \,{\scriptsize\textcolor[HTML]{228B22}{(\textbf{2.16$\times$})}} & 6.21 \,{\scriptsize\textcolor[HTML]{228B22}{(\textbf{1.12$\times$})}} & 161.01 \,{\scriptsize\textcolor[HTML]{228B22}{(\textbf{1.12$\times$})}} \\
  SparseVLM & 92.40 & 98.20 & 87.40 & 81.40 & 89.85 \,{\scriptsize\textcolor{gray}{(95.7\%)}} & 0.8040 \,{\scriptsize\textcolor[HTML]{228B22}{(\textbf{1.94$\times$})}} & 6.00 \,{\scriptsize\textcolor[HTML]{228B22}{(\textbf{1.08$\times$})}} & 166.78 \,{\scriptsize\textcolor[HTML]{228B22}{(\textbf{1.08$\times$})}} \\
  DivPrune  & 82.40 & 80.00 & 61.20 & 55.80 & 69.85 \,{\scriptsize\textcolor{gray}{(74.4\%)}} & 0.6920 \,{\scriptsize\textcolor[HTML]{228B22}{(\textbf{2.25$\times$})}} & 6.24 \,{\scriptsize\textcolor[HTML]{228B22}{(\textbf{1.13$\times$})}} & 160.23 \,{\scriptsize\textcolor[HTML]{228B22}{(\textbf{1.13$\times$})}} \\
  VLA-Cache & 97.00 & 98.80 & 92.40 & 83.80 & 93.00 \,{\scriptsize\textcolor{gray}{(99.0\%)}} & 0.8596 \,{\scriptsize\textcolor[HTML]{228B22}{(\textbf{1.81$\times$})}} & 6.04 \,{\scriptsize\textcolor[HTML]{228B22}{(\textbf{1.09$\times$})}} & 165.66 \,{\scriptsize\textcolor[HTML]{228B22}{(\textbf{1.09$\times$})}} \\
  TVCache (Ours)      & \textbf{97.80} & \textbf{99.20} & \textbf{95.00} & \textbf{87.00} & \textbf{94.75} \,{\scriptsize\textcolor{gray}{(100.9\%)}} & 0.8631 \,{\scriptsize\textcolor[HTML]{228B22}{(\textbf{1.80$\times$})}} & 6.03 \,{\scriptsize\textcolor[HTML]{228B22}{(\textbf{1.09$\times$})}} & 165.82 \,{\scriptsize\textcolor[HTML]{228B22}{(\textbf{1.09$\times$})}} \\ \midrule
  
  \rowcolor[HTML]{F2F2F2} \multicolumn{9}{c}{Retain 12.5\% Tokens} \\
  FastV     & 95.20 & 97.40 & 80.80 & 71.60 & 86.25 \,{\scriptsize\textcolor{gray}{(91.9\%)}} & 0.5852 \,{\scriptsize\textcolor[HTML]{228B22}{(\textbf{2.66$\times$})}} & 6.23 \,{\scriptsize\textcolor[HTML]{228B22}{(\textbf{1.13$\times$})}} & 160.42 \,{\scriptsize\textcolor[HTML]{228B22}{(\textbf{1.13$\times$})}} \\
  SparseVLM & 79.00 & 92.60 & 79.00 & 71.60 & 80.55 \,{\scriptsize\textcolor{gray}{(85.8\%)}} & 0.6028 \,{\scriptsize\textcolor[HTML]{228B22}{(\textbf{2.58$\times$})}} & 6.09 \,{\scriptsize\textcolor[HTML]{228B22}{(\textbf{1.10$\times$})}} & 164.21 \,{\scriptsize\textcolor[HTML]{228B22}{(\textbf{1.10$\times$})}} \\
  DivPrune  & 51.00 & 26.80 & 31.80 & 10.20 & 29.95 \,{\scriptsize\textcolor{gray}{(31.9\%)}} & 0.5721 \,{\scriptsize\textcolor[HTML]{228B22}{(\textbf{2.72$\times$})}} & 6.28 \,{\scriptsize\textcolor[HTML]{228B22}{(\textbf{1.14$\times$})}} & 159.12 \,{\scriptsize\textcolor[HTML]{228B22}{(\textbf{1.14$\times$})}} \\
  VLA-Cache & 97.00 & 98.40 & 92.00 & 82.60 & 92.50 \,{\scriptsize\textcolor{gray}{(98.5\%)}} & 0.7565 \,{\scriptsize\textcolor[HTML]{228B22}{(\textbf{2.06$\times$})}} & 6.10 \,{\scriptsize\textcolor[HTML]{228B22}{(\textbf{1.10$\times$})}} & 163.85 \,{\scriptsize\textcolor[HTML]{228B22}{(\textbf{1.10$\times$})}} \\
  TVCache (Ours)      & \textbf{97.60} & \textbf{99.20} & \textbf{94.40} & \textbf{84.40} & \textbf{93.90} \,{\scriptsize\textcolor{gray}{(100.0\%)}} & 0.7474 \,{\scriptsize\textcolor[HTML]{228B22}{(\textbf{2.08$\times$})}} & 6.11 \,{\scriptsize\textcolor[HTML]{228B22}{(\textbf{1.11$\times$})}} & 163.68 \,{\scriptsize\textcolor[HTML]{228B22}{(\textbf{1.11$\times$})}} \\ 
  
  \bottomrule
  \end{tabular}
  }
\end{table*}

\begin{table*}[t]
  \caption{Comparison of latency and performance between $\pi_{0.5}$ and VLA-Adapter evaluated on the LIBERO Benchmark.}
  \label{tab:performance_metrics_2}
  \centering
  \resizebox{\textwidth}{!}{
  \begin{tabular}{lcccccccc}
  \toprule
  Method & Spatial (\%) & Object (\%) & Goal (\%) & Long (\%) & Avg Success Rate (\%) $\uparrow$ & FLOPs (T) $\downarrow$ & Control Freq. (Hz) $\uparrow$ & Latency (ms) $\downarrow$ \\ 
  \midrule
  
  \rowcolor[HTML]{DDEBF7} 
  \multicolumn{9}{c}{\textbf{$\pi_{0.5}$}} \\ \midrule
  
  \rowcolor[HTML]{F2F2F2} \multicolumn{9}{c}{100\% Tokens} \\
  Vanilla   & 98.00 & 97.20 & 97.00 & 93.00 & 96.30 \,{\scriptsize\textcolor{gray}{(100.0\%)}} & 2.1153 \,{\scriptsize\textcolor{gray}{(1.00$\times$)}} & 8.14 \,{\scriptsize\textcolor{gray}{(1.00$\times$)}} & 122.85 \,{\scriptsize\textcolor{gray}{(1.00$\times$)}} \\ \midrule
  
  \rowcolor[HTML]{F2F2F2} \multicolumn{9}{c}{Retain 50\% Tokens} \\
  FastV     & 57.00 & 67.20 & 60.80 & 59.40 & 61.10 \,{\scriptsize\textcolor{gray}{(63.4\%)}} & 1.6061 \,{\scriptsize\textcolor[HTML]{228B22}{(\textbf{1.32$\times$})}} & 8.90 \,{\scriptsize\textcolor[HTML]{228B22}{(\textbf{1.09$\times$})}} & 112.34 \,{\scriptsize\textcolor[HTML]{228B22}{(\textbf{1.09$\times$})}} \\
  SparseVLM & 78.20 & 73.00 & 86.40 & 43.60 & 70.30 \,{\scriptsize\textcolor{gray}{(73.0\%)}} & 1.5429 \,{\scriptsize\textcolor[HTML]{228B22}{(\textbf{1.37$\times$})}} & 8.82 \,{\scriptsize\textcolor[HTML]{228B22}{(\textbf{1.08$\times$})}} & 113.38 \,{\scriptsize\textcolor[HTML]{228B22}{(\textbf{1.08$\times$})}} \\
  DivPrune  & 88.80 & 83.80 & 88.60 & 68.20 & 82.35 \,{\scriptsize\textcolor{gray}{(85.5\%)}} & 1.5425 \,{\scriptsize\textcolor[HTML]{228B22}{(\textbf{1.37$\times$})}} & 8.94 \,{\scriptsize\textcolor[HTML]{228B22}{(\textbf{1.10$\times$})}} & 111.83 \,{\scriptsize\textcolor[HTML]{228B22}{(\textbf{1.10$\times$})}} \\
  VLA-Cache & 96.20 & 97.60 & 96.80 & 86.20 & 94.20 \,{\scriptsize\textcolor{gray}{(97.8\%)}} & 1.7206 \,{\scriptsize\textcolor[HTML]{228B22}{(\textbf{1.23$\times$})}} & 8.85 \,{\scriptsize\textcolor[HTML]{228B22}{(\textbf{1.09$\times$})}} & 113.04 \,{\scriptsize\textcolor[HTML]{228B22}{(\textbf{1.09$\times$})}} \\
  TVCache (Ours)      & \textbf{97.80} & \textbf{98.00} & \textbf{98.00} & \textbf{89.60} & \textbf{95.85} \,{\scriptsize\textcolor{gray}{(99.5\%)}} & 1.6374 \,{\scriptsize\textcolor[HTML]{228B22}{(\textbf{1.29$\times$})}} & 8.93 \,{\scriptsize\textcolor[HTML]{228B22}{(\textbf{1.10$\times$})}} & 112.02 \,{\scriptsize\textcolor[HTML]{228B22}{(\textbf{1.10$\times$})}} \\ \midrule
  
  \rowcolor[HTML]{F2F2F2} \multicolumn{9}{c}{Retain 25\% Tokens} \\
  FastV     & 24.40 & 18.60 & 25.40 & 7.20 & 18.90 \,{\scriptsize\textcolor{gray}{(19.6\%)}} & 1.3548 \,{\scriptsize\textcolor[HTML]{228B22}{(\textbf{1.56$\times$})}} & 9.47 \,{\scriptsize\textcolor[HTML]{228B22}{(\textbf{1.16$\times$})}} & 105.63 \,{\scriptsize\textcolor[HTML]{228B22}{(\textbf{1.16$\times$})}} \\
  SparseVLM & 35.60 & 24.00 & 50.00 & 5.20 & 28.70 \,{\scriptsize\textcolor{gray}{(29.8\%)}} & 1.2613 \,{\scriptsize\textcolor[HTML]{228B22}{(\textbf{1.68$\times$})}} & 9.35 \,{\scriptsize\textcolor[HTML]{228B22}{(\textbf{1.15$\times$})}} & 106.99 \,{\scriptsize\textcolor[HTML]{228B22}{(\textbf{1.15$\times$})}} \\
  DivPrune  & 46.40 & 48.40 & 69.40 & 38.00 & 50.55 \,{\scriptsize\textcolor{gray}{(52.5\%)}} & 1.2597 \,{\scriptsize\textcolor[HTML]{228B22}{(\textbf{1.68$\times$})}} & 9.58 \,{\scriptsize\textcolor[HTML]{228B22}{(\textbf{1.18$\times$})}} & 104.34 \,{\scriptsize\textcolor[HTML]{228B22}{(\textbf{1.18$\times$})}} \\
  VLA-Cache & 86.00 & 92.80 & 91.40 & 77.20 & 86.85 \,{\scriptsize\textcolor{gray}{(90.2\%)}} & 1.6083 \,{\scriptsize\textcolor[HTML]{228B22}{(\textbf{1.32$\times$})}} & 9.36 \,{\scriptsize\textcolor[HTML]{228B22}{(\textbf{1.15$\times$})}} & 106.81 \,{\scriptsize\textcolor[HTML]{228B22}{(\textbf{1.15$\times$})}} \\
  TVCache (Ours)      & \textbf{93.80} & \textbf{97.80} & \textbf{96.60} & \textbf{87.40} & \textbf{93.90} \,{\scriptsize\textcolor{gray}{(97.5\%)}} & 1.5109 \,{\scriptsize\textcolor[HTML]{228B22}{(\textbf{1.40$\times$})}} & 9.38 \,{\scriptsize\textcolor[HTML]{228B22}{(\textbf{1.15$\times$})}} & 106.63 \,{\scriptsize\textcolor[HTML]{228B22}{(\textbf{1.15$\times$})}} \\ 
  
  \midrule\midrule
  
  \rowcolor[HTML]{DDEBF7} 
  \multicolumn{9}{c}{\textbf{VLA-Adapter}} \\ \midrule
  
  \rowcolor[HTML]{F2F2F2} \multicolumn{9}{c}{100\% Tokens} \\
  Vanilla   & 98.40 & 96.80 & 96.80 & 94.80 & 96.70 \,{\scriptsize\textcolor{gray}{(100.0\%)}} & 3.9732 \,{\scriptsize\textcolor{gray}{(1.00$\times$)}} & 10.71 \,{\scriptsize\textcolor{gray}{(1.00$\times$)}} & 93.40 \,{\scriptsize\textcolor{gray}{(1.00$\times$)}} \\ \midrule
  
  \rowcolor[HTML]{F2F2F2} \multicolumn{9}{c}{Retain 50\% Tokens} \\
  VLA-Cache & 88.60 & 69.40 & 77.80 & 73.60 & 77.35 \,{\scriptsize\textcolor{gray}{(80.0\%)}} & 2.6821 \,{\scriptsize\textcolor[HTML]{228B22}{(\textbf{1.48$\times$})}} & 12.60 \,{\scriptsize\textcolor[HTML]{228B22}{(\textbf{1.18$\times$})}} & 79.36 \,{\scriptsize\textcolor[HTML]{228B22}{(\textbf{1.18$\times$})}} \\
  TVCache (Ours)      & \textbf{91.60} & \textbf{73.20} & \textbf{78.80} & \textbf{76.20} & \textbf{79.95} \,{\scriptsize\textcolor{gray}{(82.7\%)}} & 2.6713 \,{\scriptsize\textcolor[HTML]{228B22}{(\textbf{1.49$\times$})}} & 12.72 \,{\scriptsize\textcolor[HTML]{228B22}{(\textbf{1.19$\times$})}} & 78.59 \,{\scriptsize\textcolor[HTML]{228B22}{(\textbf{1.19$\times$})}} \\ \midrule
  
  \rowcolor[HTML]{F2F2F2} \multicolumn{9}{c}{Retain 25\% Tokens} \\
  VLA-Cache & 55.60 & 44.40 & 43.40 & 50.00 & 48.35 \,{\scriptsize\textcolor{gray}{(50.0\%)}} & 2.2305 \,{\scriptsize\textcolor[HTML]{228B22}{(\textbf{1.78$\times$})}} & 14.22 \,{\scriptsize\textcolor[HTML]{228B22}{(\textbf{1.33$\times$})}} & 70.33 \,{\scriptsize\textcolor[HTML]{228B22}{(\textbf{1.33$\times$})}} \\
  TVCache (Ours)      & \textbf{67.20} & \textbf{47.20} & \textbf{47.60} & \textbf{55.60} & \textbf{54.40} \,{\scriptsize\textcolor{gray}{(56.3\%)}} & 2.2082 \,{\scriptsize\textcolor[HTML]{228B22}{(\textbf{1.80$\times$})}} & 14.61 \,{\scriptsize\textcolor[HTML]{228B22}{(\textbf{1.36$\times$})}} & 68.46 \,{\scriptsize\textcolor[HTML]{228B22}{(\textbf{1.36$\times$})}} \\ 
  
  \bottomrule
  \end{tabular}
  }
\end{table*}

\begin{table*}[t]
  \caption{Comparison of performance for VLA-Adapter evaluated on the CALVIN Benchmark.}
  \label{tab:performance_metrics_3}
  \centering
  \resizebox{\textwidth}{!}{
  \begin{tabular}{lccccccccc}
  \toprule
  Method & Task 1 (\%) & Task 2 (\%) & Task 3 (\%) & Task 4 (\%) & Task 5 (\%) & Avg Success Rate (\%) $\uparrow$ & FLOPs (T) $\downarrow$ & Control Freq. (Hz) $\uparrow$ & Latency (ms) $\downarrow$ \\ 
  \midrule
  
  \rowcolor[HTML]{DDEBF7} 
  \multicolumn{10}{c}{\textbf{VLA-Adapter}} \\ \midrule
  
  \rowcolor[HTML]{F2F2F2} \multicolumn{10}{c}{100\% Tokens} \\
  Vanilla   & 98.50 & 94.60 & 89.20 & 84.90 & 78.90 & 89.22 \,{\scriptsize\textcolor{gray}{(100.0\%)}} & 0.2597 \,{\scriptsize\textcolor{gray}{(1.00$\times$)}} & 21.45 \,{\scriptsize\textcolor{gray}{(1.00$\times$)}} & 46.62 \,{\scriptsize\textcolor{gray}{(1.00$\times$)}} \\ \midrule
  
  \rowcolor[HTML]{F2F2F2} \multicolumn{10}{c}{Retain 50\% Tokens} \\
  VLA-Cache & 95.60 & 91.00 & 84.30 & 76.80 & 69.40 & 83.42 \,{\scriptsize\textcolor{gray}{(93.5\%)}} & 0.1831 \,{\scriptsize\textcolor[HTML]{228B22}{(\textbf{1.42$\times$})}} & 22.72 \,{\scriptsize\textcolor[HTML]{228B22}{(\textbf{1.06$\times$})}} & 44.01 \,{\scriptsize\textcolor[HTML]{228B22}{(\textbf{1.06$\times$})}} \\
  TVCache (Ours)      & \textbf{97.10} & \textbf{92.50} & \textbf{87.90} & \textbf{81.10} & \textbf{73.10} & \textbf{86.34} \,{\scriptsize\textcolor{gray}{(96.8\%)}} & 0.1751 \,{\scriptsize\textcolor[HTML]{228B22}{(\textbf{1.48$\times$})}} & 22.73 \,{\scriptsize\textcolor[HTML]{228B22}{(\textbf{1.06$\times$})}} & 43.99 \,{\scriptsize\textcolor[HTML]{228B22}{(\textbf{1.06$\times$})}} \\ \midrule
  
  \rowcolor[HTML]{F2F2F2} \multicolumn{10}{c}{Retain 25\% Tokens} \\
  VLA-Cache & 75.80 & 61.00 & 44.80 & 34.00 & 24.70 & 48.06 \,{\scriptsize\textcolor{gray}{(53.9\%)}} & 0.1472 \,{\scriptsize\textcolor[HTML]{228B22}{(\textbf{1.76$\times$})}} & 22.82 \,{\scriptsize\textcolor[HTML]{228B22}{(\textbf{1.06$\times$})}} & 43.82 \,{\scriptsize\textcolor[HTML]{228B22}{(\textbf{1.06$\times$})}} \\
  TVCache (Ours)      & \textbf{82.70} & \textbf{69.10} & \textbf{56.70} & \textbf{46.40} & \textbf{36.30} & \textbf{58.24} \,{\scriptsize\textcolor{gray}{(65.3\%)}} & 0.1351 \,{\scriptsize\textcolor[HTML]{228B22}{(\textbf{1.92$\times$})}} & 22.87 \,{\scriptsize\textcolor[HTML]{228B22}{(\textbf{1.07$\times$})}} & 43.72 \,{\scriptsize\textcolor[HTML]{228B22}{(\textbf{1.07$\times$})}} \\ 
  
  \bottomrule
  \end{tabular}
  }
\end{table*}

\begin{table}[t]
    \caption{Descriptions of the evaluation tasks.}
    \label{tab:real_world_tasks}
    \centering
    \renewcommand{\arraystretch}{0.95}
    \setlength{\tabcolsep}{4pt}
    
    \begin{tabularx}{\columnwidth}{
        l >{\raggedright\arraybackslash}X
    }
    \toprule
    \textbf{Task Name} & \textbf{Description} \\ 
    \midrule
    \texttt{SetKettle} & Pick up the kettle and place it on the plate. \\
    \texttt{PackPear}  & Pick up the pear and place it inside the basket. \\
    \texttt{HangCup}   & Hang the cup on the wooden mug tree. \\
    \texttt{PotDuck}   & Put the duck into the pot and cover it with the lid. \\ 
    \bottomrule
    \end{tabularx}
    \vspace{-0.3cm}
\end{table}

\section{Experiment}
\label{sec:result}
We evaluate TVCache across simulated and real-world environments using four VLA architectures: OpenVLA-OFT \cite{kim2025fine}, BitVLA \cite{wang2025bitvla}, $\pi_{0.5}$ \cite{intelligence2025pi_}, and VLA-Adapter \cite{wang2025vla}.
Real-world robustness is validated by deploying OpenVLA-OFT on a physical robotic platform.
All experiments are executed on a workstation with four NVIDIA RTX 4090 GPUs.

\subsection{Experimental Setup}
\textbf{Baselines.} 
We compare TVCache with a comprehensive suite of training-free accelerators under given compute budgets, including FastV \cite{chen2024image}, SparseVLM \cite{zhang2024sparsevlm}, DivPrune \cite{alvar2025divprune}, and the VLA-specific VLA-Cache \cite{xu2025vla}. For TVCache, we fix $\alpha=\beta=0.5$, $K=4$, and $d_{\min}=2$ across all models and benchmarks, with $\gamma^{(l)}=[0.5,0.65,0.8,1.0]$.

\textbf{Efficiency Profiling.}
All latency measurements are conducted on a single NVIDIA RTX 4090 with batch size 1 using BF16 precision. We exclude the first 10 warm-up runs, synchronize CUDA before and after each timed query, and report the mean over 500 measured runs.

\textbf{Evaluation Metrics.}
To ensure a fair and rigorous benchmark, all methods are evaluated under identical token reduction constraints. We systematically investigate performance across varying retention ratios (50\%, 25\%, and 12.5\%). Following established evaluation protocols in prior research on efficient multimodal architectures \cite{chen2024image, xu2025vla, liu2025vla}, efficacy is quantified using four metrics: (1) task success rate (\%); (2) computational cost of action prediction (FLOPs, T); (3) control frequency (Hz); and (4) end-to-end inference latency (ms). Control frequency is computed as the inverse of end-to-end latency, reflecting how frequently the policy can update its action prediction during continuous robot control.

\textbf{Evaluation Benchmark.}
We evaluate TVCache across simulated and physical environments. \textbf{LIBERO} \cite{liu2023libero}: We test OpenVLA-OFT, BitVLA, $\pi_{0.5}$, and VLA-Adapter across four task suites, totaling 500 evaluation episodes per suite under official protocols. \textbf{CALVIN} \cite{mees2022calvin}: We assess the VLA-Adapter's generalization in the challenging \textit{Split ABC $\rightarrow$ D} setting, evaluating the model in a novel environment after training on prior domains. \textbf{Real Robot Evaluation}: We deploy a LoRA-finetuned OpenVLA-OFT \cite{hu2022lora} on a Franka Research 3 manipulator \cite{haddadin2022franka} equipped with dual cameras. Performance is benchmarked across four manipulation tasks summarized in Table~\ref{tab:real_world_tasks}, using 100 teleoperated demonstrations and 50 evaluation trials per task.

\subsection{Results and Analysis}
\textbf{LIBERO Benchmark.}
As shown in Table~\ref{tab:performance_metrics_1}, TVCache consistently outperforms VLA-Cache on OpenVLA-OFT and BitVLA, with larger gains under aggressive token reuse. On OpenVLA-OFT, TVCache achieves 97.55\% and 95.55\% success at 50\% and 25\% retention, respectively. At 12.5\% retention, it improves success from 69.50\% to 84.00\% with $2.45\times$ lower FLOPs than full-token inference. Meanwhile, control frequency increases from 10.43\,Hz to 15.04\,Hz and latency decreases from 95.87\,ms to 66.51\,ms.
On BitVLA, TVCache achieves 94.90\%, 94.75\%, and 93.90\% success at 50\%, 25\%, and 12.5\% retention, respectively, matching or exceeding the 93.90\% full-token baseline. 

Table~\ref{tab:performance_metrics_2} further demonstrates the generality of TVCache on $\pi_{0.5}$ and VLA-Adapter. At 25\% retention, TVCache improves over VLA-Cache from 86.85\% to 93.90\% on $\pi_{0.5}$ and from 48.35\% to 54.40\% on VLA-Adapter. Similar gains are also observed at 50\% retention, where TVCache improves the success rate from 94.20\% to 95.85\% on $\pi_{0.5}$ and from 77.35\% to 79.95\% on VLA-Adapter. These results demonstrate TVCache's generality across different VLA architectures, with larger gains under tighter token budgets.

\textbf{CALVIN Benchmark.}
As shown in Table~\ref{tab:performance_metrics_3}, TVCache also generalizes to the challenging \textit{Split ABC $\rightarrow$ D} setting. At 25\% retention, it improves the success rate from 48.06\% to 58.24\% over VLA-Cache while achieving a $1.92\times$ FLOPs reduction relative to full-token inference. The latency gain is more modest, reducing latency from 46.62\,ms to 43.72\,ms and increasing control frequency from 21.45\,Hz to 22.87\,Hz, as the lightweight VLA-Adapter makes routing overhead relatively more significant.

\begin{table}[t]
\caption{Runtime overhead breakdown of TVCache under different token retention ratios.}
\label{tab:runtime_overhead}
\centering
\scriptsize
\resizebox{\columnwidth}{!}{%
\begin{tabular}{lllllll}
\toprule
TVCache Ret. & Patch sim. & Head filtering & Token selection &
Reuse-layer selection & Bookkeeping & Total (ms) \\
\midrule
50\%   & 0.742 & 0.746 & 0.322 & 1.012 & 0.066 & 2.888 \\
25\%   & 0.651 & 0.747 & 0.316 & 1.052 & 0.073 & 2.839 \\
12.5\% & 0.782 & 0.708 & 0.304 & 0.968 & 0.075 & 2.837 \\
\bottomrule
\end{tabular}
}
\vspace{-0.3cm}
\end{table}

As shown in Table~\ref{tab:runtime_overhead}, TVCache introduces only 2.84--2.89\,ms of nearly constant runtime overhead. Reuse-layer selection is the main contributor, while token selection and bookkeeping add less than 0.4\,ms. This fixed cost is more noticeable on lightweight models and less significant on larger VLA backbones.

\begin{figure}[t]
    \centering
    \includegraphics[
        width=\columnwidth,
        trim={0 0 0 0},
        clip
    ]{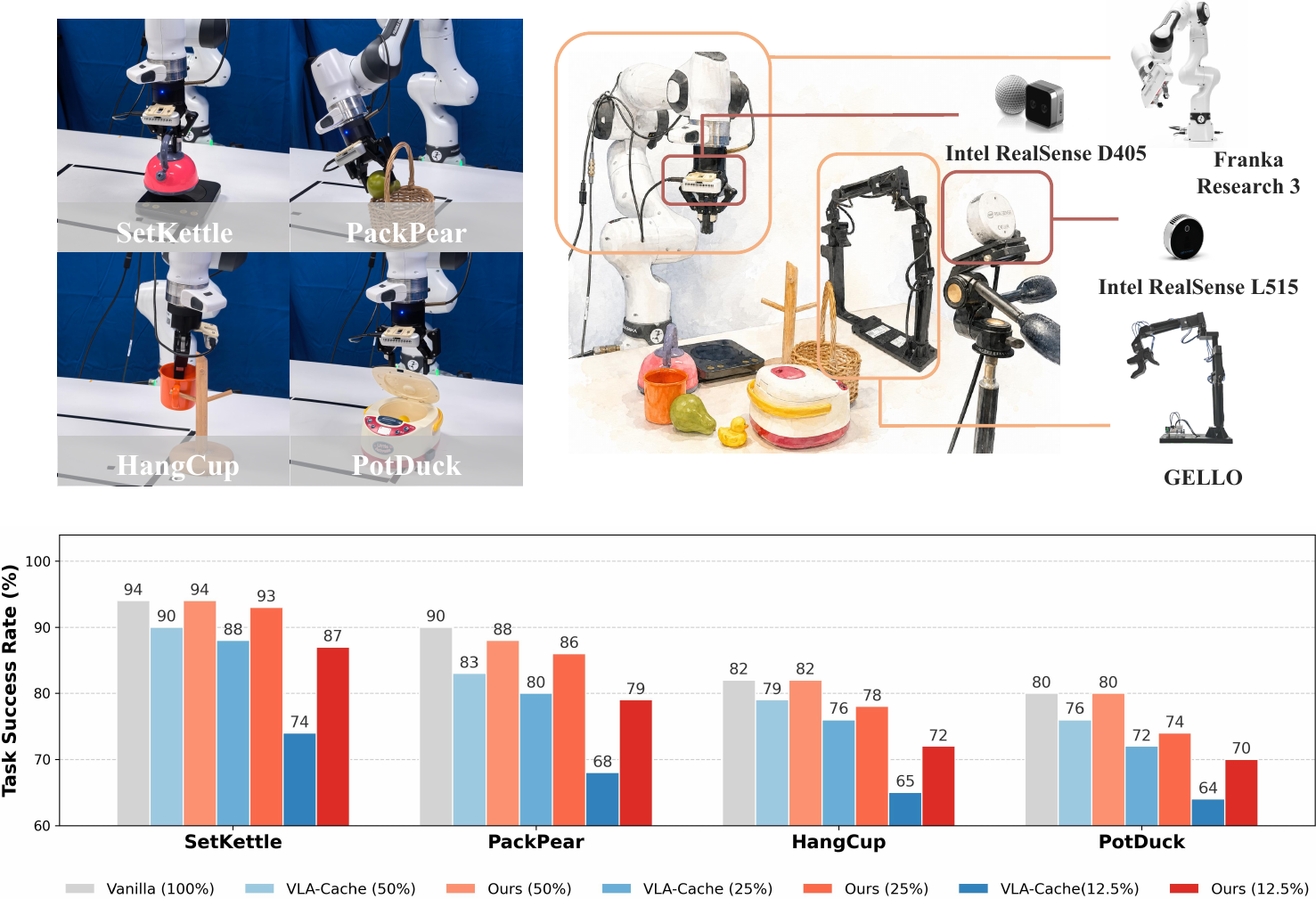}
    \caption{Real-world robot evaluation across four manipulation tasks. The figure shows the physical task scenarios and the corresponding success rates of TVCache and VLA-Cache under different token retention ratios.}
    \label{fig:flat_bar_chart}
    \vspace{-0.3cm}
\end{figure}

\begin{table}[t]
\caption{Ablation study on OpenVLA-OFT and BitVLA.}
\label{tab:ablation_study}
\centering

\renewcommand{\arraystretch}{0.92}
\setlength{\tabcolsep}{2.5pt}
\scriptsize

\resizebox{\columnwidth}{!}{%
\begin{tabular}{lccccc}
\toprule
Method & Spatial & Object & Goal & Long & Avg. $\uparrow$ \\ 
\midrule

\multicolumn{6}{c}{\textbf{OpenVLA-OFT}} \\ 
\midrule

\rowcolor[HTML]{F2F2F2} 
\multicolumn{6}{c}{100\% Tokens} \\ 
Vanilla
    & 99.00 & 98.00 & 97.20 & 94.40 & 97.15 \\ 
\midrule

\rowcolor[HTML]{F2F2F2} 
\multicolumn{6}{c}{Retain 50\% Tokens} \\ 
VLA-Cache
    & 98.20 & 98.20 & 97.40 & 93.40 & 96.80 \\
\quad + Head
    & 98.60 {\tiny($\uparrow0.40$)}
    & 98.40 {\tiny($\uparrow0.20$)}
    & 97.80 {\tiny($\uparrow0.40$)}
    & 93.80 {\tiny($\uparrow0.40$)}
    & 97.15 {\tiny($\uparrow0.35$)} \\
\quad + Layer
    & 99.00 {\tiny($\uparrow0.80$)}
    & 98.20 {\tiny($-$)}
    & 97.20 {\tiny($\downarrow0.20$)}
    & 93.40 {\tiny($-$)}
    & 96.95 {\tiny($\uparrow0.15$)} \\
Ours
    & \textbf{99.20} {\tiny($\uparrow1.00$)}
    & \textbf{98.80} {\tiny($\uparrow0.60$)}
    & \textbf{97.80} {\tiny($\uparrow0.40$)}
    & \textbf{94.40} {\tiny($\uparrow1.00$)}
    & \textbf{97.55} {\tiny($\uparrow0.75$)} \\ 
\midrule

\rowcolor[HTML]{F2F2F2}
\multicolumn{6}{c}{Retain 25\% Tokens} \\ 
VLA-Cache
    & 96.40 & 84.40 & 94.00 & 76.60 & 87.85 \\
\quad + Head
    & 96.20 {\tiny($\downarrow0.20$)}
    & 94.00 {\tiny($\uparrow9.60$)}
    & 96.80 {\tiny($\uparrow2.80$)}
    & 81.20 {\tiny($\uparrow4.60$)}
    & 92.05 {\tiny($\uparrow4.20$)} \\
\quad + Layer
    & 97.40 {\tiny($\uparrow1.00$)}
    & 92.40 {\tiny($\uparrow8.00$)}
    & 96.20 {\tiny($\uparrow2.20$)}
    & 83.40 {\tiny($\uparrow6.80$)}
    & 92.35 {\tiny($\uparrow4.50$)} \\
Ours
    & \textbf{98.20} {\tiny($\uparrow1.80$)}
    & \textbf{96.00} {\tiny($\uparrow11.60$)}
    & \textbf{97.80} {\tiny($\uparrow3.80$)}
    & \textbf{90.20} {\tiny($\uparrow13.60$)}
    & \textbf{95.55} {\tiny($\uparrow7.70$)} \\ 
\midrule

\rowcolor[HTML]{F2F2F2}
\multicolumn{6}{c}{Retain 12.5\% Tokens} \\ 
VLA-Cache
    & 83.60 & 62.00 & 82.80 & 49.60 & 69.50 \\
\quad + Head
    & 85.80 {\tiny($\uparrow2.20$)}
    & 65.40 {\tiny($\uparrow3.40$)}
    & 82.60 {\tiny($\downarrow0.20$)}
    & 49.80 {\tiny($\uparrow0.20$)}
    & 70.90 {\tiny($\uparrow1.40$)} \\
\quad + Layer
    & 94.60 {\tiny($\uparrow11.00$)}
    & 83.60 {\tiny($\uparrow21.60$)}
    & 93.00 {\tiny($\uparrow10.20$)}
    & 57.60 {\tiny($\uparrow8.00$)}
    & 82.20 {\tiny($\uparrow12.70$)} \\
Ours
    & \textbf{96.80} {\tiny($\uparrow13.20$)}
    & \textbf{84.80} {\tiny($\uparrow22.80$)}
    & \textbf{94.40} {\tiny($\uparrow11.60$)}
    & \textbf{60.00} {\tiny($\uparrow10.40$)}
    & \textbf{84.00} {\tiny($\uparrow14.50$)} \\ 
    
\midrule\midrule

\multicolumn{6}{c}{\textbf{BitVLA}} \\ 
\midrule

\rowcolor[HTML]{F2F2F2}
\multicolumn{6}{c}{100\% Tokens} \\ 
Vanilla
    & 97.60 & 99.40 & 91.40 & 87.20 & 93.90 \\ 
\midrule

\rowcolor[HTML]{F2F2F2}
\multicolumn{6}{c}{Retain 50\% Tokens} \\ 
VLA-Cache
    & 97.20 & 98.80 & 93.60 & 86.00 & 93.90 \\
\quad + Head
    & 97.40 {\tiny($\uparrow0.20$)}
    & 99.00 {\tiny($\uparrow0.20$)}
    & 92.80 {\tiny($\downarrow0.80$)}
    & 87.20 {\tiny($\uparrow1.20$)}
    & 94.10 {\tiny($\uparrow0.20$)} \\
\quad + Layer
    & 97.80 {\tiny($\uparrow0.60$)}
    & 99.20 {\tiny($\uparrow0.40$)}
    & 93.00 {\tiny($\downarrow0.60$)}
    & 86.60 {\tiny($\uparrow0.60$)}
    & 94.15 {\tiny($\uparrow0.25$)} \\
Ours
    & \textbf{98.00} {\tiny($\uparrow0.80$)}
    & \textbf{99.40} {\tiny($\uparrow0.60$)}
    & \textbf{95.00} {\tiny($\uparrow1.40$)}
    & \textbf{87.20} {\tiny($\uparrow1.20$)}
    & \textbf{94.90} {\tiny($\uparrow1.00$)} \\ 
\midrule

\rowcolor[HTML]{F2F2F2}
\multicolumn{6}{c}{Retain 25\% Tokens} \\ 
VLA-Cache
    & 97.00 & 98.80 & 92.40 & 83.80 & 93.00 \\
\quad + Head
    & 97.00 {\tiny($-$)}
    & 99.20 {\tiny($\uparrow0.40$)}
    & 93.40 {\tiny($\uparrow1.00$)}
    & 84.40 {\tiny($\uparrow0.60$)}
    & 93.50 {\tiny($\uparrow0.50$)} \\
\quad + Layer
    & 97.60 {\tiny($\uparrow0.60$)}
    & 99.20 {\tiny($\uparrow0.40$)}
    & 93.20 {\tiny($\uparrow0.80$)}
    & 85.40 {\tiny($\uparrow1.60$)}
    & 93.85 {\tiny($\uparrow0.85$)} \\
Ours
    & \textbf{97.80} {\tiny($\uparrow0.80$)}
    & \textbf{99.20} {\tiny($\uparrow0.40$)}
    & \textbf{95.00} {\tiny($\uparrow2.60$)}
    & \textbf{87.00} {\tiny($\uparrow3.20$)}
    & \textbf{94.75} {\tiny($\uparrow1.75$)} \\ 
\midrule

\rowcolor[HTML]{F2F2F2}
\multicolumn{6}{c}{Retain 12.5\% Tokens} \\ 
VLA-Cache
    & 97.00 & 98.40 & 92.00 & 82.60 & 92.50 \\
\quad + Head
    & 97.40 {\tiny($\uparrow0.40$)}
    & 99.20 {\tiny($\uparrow0.80$)}
    & 94.60 {\tiny($\uparrow2.60$)}
    & 81.40 {\tiny($\downarrow1.20$)}
    & 93.15 {\tiny($\uparrow0.65$)} \\
\quad + Layer
    & 97.20 {\tiny($\uparrow0.20$)}
    & 99.00 {\tiny($\uparrow0.60$)}
    & 93.60 {\tiny($\uparrow1.60$)}
    & 83.60 {\tiny($\uparrow1.00$)}
    & 93.35 {\tiny($\uparrow0.85$)} \\
Ours
    & \textbf{97.60} {\tiny($\uparrow0.60$)}
    & \textbf{99.20} {\tiny($\uparrow0.80$)}
    & \textbf{94.40} {\tiny($\uparrow2.40$)}
    & \textbf{84.40} {\tiny($\uparrow1.80$)}
    & \textbf{93.90} {\tiny($\uparrow1.40$)} \\ 
\bottomrule
\end{tabular}%
}
\vspace{-0.3cm}
\end{table}

\begin{figure}[t]
    \centering
    \includegraphics[
        width=\columnwidth,
        trim={0 0 0 0},
        clip
    ]{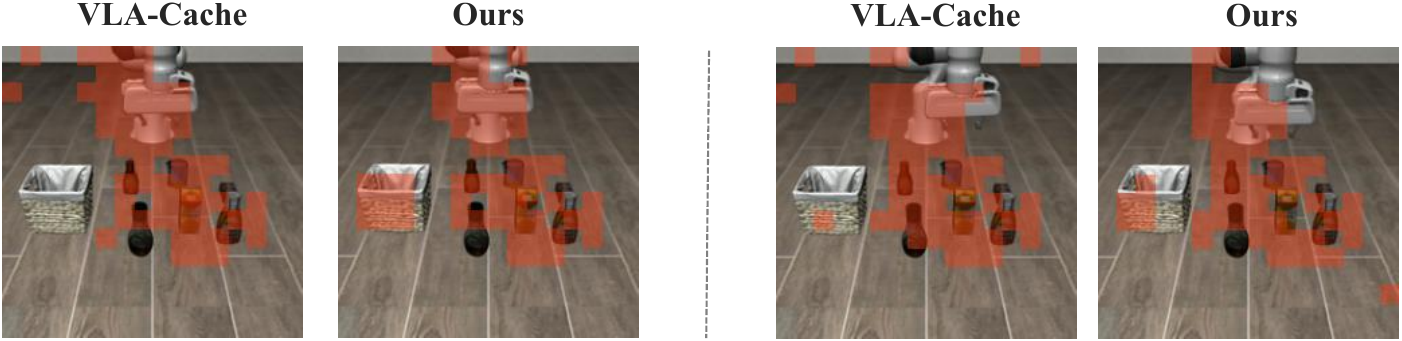}
    \caption{Qualitative comparison of cross-modal attention. TVCache filters degraded heads, reducing background noise and spatial misalignment for more task-relevant visual grounding.}
    \label{fig:attention_vis}
    \vspace{-0.3cm}
\end{figure}

\textbf{Real Robot Deployment.}
As shown in Fig.~\ref{fig:flat_bar_chart}, TVCache consistently surpasses VLA-Cache across all retention ratios in physical deployments. At 50\% retention, it achieves near-lossless performance, matching the full-token baseline on tasks like SetKettle (94\%) and HangCup (82\%). Crucially, under extreme 12.5\% retention, TVCache maintains a robust 87\% task success rate on SetKettle, demonstrating superior robustness and control precision for real-world manipulation.

\textbf{Ablation Study.}
Ablation results in Table~\ref{tab:ablation_study} validate the contributions of both attention-head filtering and reuse-layer selection. Attention-head filtering alleviates token-level spatial misalignment by emphasizing task-relevant visual information, while reuse-layer selection reduces the impact of unstable layer reuse.
The benefits become particularly pronounced under aggressive token reuse. On OpenVLA-OFT at 12.5\% retention, VLA-Cache achieves only 69.50\% average success, while reuse-layer selection alone improves it to 82.20\%. Combining both components further raises performance to 84.00\% ($\uparrow14.50$ pp). The largest improvement occurs on the \textit{Object} suite, where TVCache increases success from 62.00\% to 84.80\% ($\uparrow22.80$ pp). Consistent improvements on BitVLA further demonstrate that both components remain effective across different VLA architectures.


\textbf{Qualitative Analysis of Attention Distributions.}
Fig.~\ref{fig:attention_vis} compares cross-modal attention distributions. VLA-Cache shows diffuse background responses and spatial misalignment, while TVCache filters degraded heads to produce more concentrated attention on task-relevant regions, improving visual grounding for action generation.

\section{Complexity Analysis}
\label{sec:complexity_analysis}

\subsection{Definitions and Baseline Complexity}
Let $D$ denote the hidden dimension, $L=L_t+L_v+L_a$ the total sequence length, and $L_r=|P_{\text{reuse}}|$ the number of reused visual tokens. We analyze only the attention-related computation affected by token reuse, excluding FFN, vision encoding, action decoding, and routing overhead. For a standard Transformer layer, the cost is approximated as:
\begin{equation}
\mathcal{C}_{\text{base}} \approx 4LD^2 + 2L^2D,
\end{equation}
where $4LD^2$ accounts for the Q, K, V, and output projections, while $2L^2D$ corresponds to attention computation.

\subsection{TVCache Computational Savings}
At selected reuse layers, the $L_r$ reused tokens reuse cached K/V states, avoiding redundant projections and query-side attention computation. The per-layer FLOPs reduction is approximated as:
\begin{equation}
\mathcal{S}_{\text{cache}}
\approx
4L_rD^2 + 2LL_rD.
\end{equation}
The first term captures projection savings, while the second accounts for reduced attention computation.

\subsection{Total Computational Complexity}
The per-layer complexity of TVCache is obtained by subtracting the caching savings from the baseline cost:
\begin{equation}
\mathcal{C}_{\text{TVCache}}
\approx
\mathcal{C}_{\text{base}}
-
\mathcal{S}_{\text{cache}}.
\end{equation}

Substituting the above terms gives:
\begin{equation}
\mathcal{C}_{\text{TVCache}}
\approx
\left(4LD^2 + 2L^2D\right)
-
\left(4L_rD^2 + 2LL_rD\right).
\end{equation}

This analysis shows that TVCache reduces projection and attention costs through token reuse, while unchanged model components and routing overhead limit the corresponding end-to-end speedup. At 12.5\% retention on OpenVLA-OFT, it achieves a $2.45\times$ FLOPs reduction and a $1.44\times$ latency speedup, while the routing overhead remains nearly constant at only 2.84--2.89\,ms across different retention ratios.

\section{Conclusion}
\label{sec:conclusion}

We present TVCache, a training-free and plug-and-play token-caching framework that exploits text-vision synergy for efficient VLA inference. TVCache addresses two limitations of existing caching methods: token-level spatial misalignment and layer-level depth mismatch. Specifically, attention-head filtering improves task-relevant visual grounding by suppressing unreliable cross-modal responses, while entropy-guided reuse-layer selection avoids unstable caching locations and enables more effective cache allocation. Experiments across four VLA architectures, two simulation benchmarks, and real-world robotic tasks consistently show improved task success under matched token-retention ratios. These results demonstrate the effectiveness of text-vision-aware cache allocation for training-free VLA acceleration.

\bibliographystyle{IEEEtran}
\bibliography{example}  

\end{document}